\documentclass[sn-apa]{sn-jnl}

\usepackage{graphicx}%
\usepackage{multirow}%
\usepackage{amsmath,amssymb,amsfonts}%
\usepackage{amsthm}%
\usepackage{mathrsfs}%
\usepackage[title]{appendix}%
\usepackage{xcolor}%
\usepackage{textcomp}%
\usepackage{manyfoot}%
\usepackage{booktabs,tabularx}%
\usepackage{algorithm}%
\usepackage{algorithmicx}%
\usepackage{algpseudocode}%
\usepackage{listings}%

\usepackage{array}
\usepackage{placeins}

\usepackage{booktabs}
\usepackage{longtable}
\usepackage{array}
\usepackage{multirow}
\usepackage{wrapfig}
\usepackage{float}
\usepackage{colortbl}
\usepackage{pdflscape}
\usepackage{tabu}
\usepackage{threeparttable}
\usepackage{threeparttablex}
\usepackage[normalem]{ulem}
\usepackage{makecell}
\usepackage{xcolor}

\newcolumntype{C}[1]{>{\centering\let\newline\\\arraybackslash\hspace{0pt}}m{#1}}

\theoremstyle{thmstyleone}%
\theoremstyle{thmstyletwo}%

\theoremstyle{thmstylethree}%

\begin{document}

\title[Article Title]{Conformal Prediction in Multi-User Settings: An Evaluation}


\author*[1]{\fnm{Enrique} \sur{Garcia-Ceja}}\email{enrique.gc@tec.mx}

\author[1]{\fnm{Luciano} \sur{Garcia-Banuelos}}\email{luciano.garcia@tec.mx}

\author[2]{\fnm{Nicolas} \sur{Jourdan}}\email{N.Jourdan@ptw.tu-darmstadt.de}


\affil*[1]{\orgname{Tecnologico de Monterrey, School of Engineering and Sciences}, \orgaddress{\street{Ave. Eugenio Garza Sada 2501}, \city{Monterrey}, \postcode{64849}, \state{N.L.}, \country{México}}}

\affil[2]{\orgname{Technical University of Darmstadt}, \orgaddress{\street{Otto-Berndt-Strasse 2}, \city{Darmstadt}, \postcode{64287}, \country{Germany}}}


\abstract{Typically, machine learning models are trained and evaluated without making any distinction between users (e.g, using traditional hold-out and cross-validation). However, this produces inaccurate performance metrics estimates in multi-user settings. That is, situations where the data were collected by multiple users with different characteristics (e.g., age, gender, height, etc.) which is very common in user computer interaction and medical applications. For these types of scenarios model evaluation strategies that provide better performance estimates have been proposed such as \emph{mixed}, \emph{user-independent}, \emph{user-dependent}, and \emph{user-adaptive} models. Although those strategies are better suited for multi-user systems, they are typically assessed with respect to performance metrics that capture the \emph{overall} behavior of the models and do not provide any performance guarantees for individual predictions nor they provide any feedback about the predictions' uncertainty. In order to overcome those limitations, in this work we evaluated the conformal prediction framework in several multi-user settings. Conformal prediction is a model agnostic method that provides confidence guarantees on the predictions, thus, increasing the trustworthiness and robustness of the models. We conducted extensive experiments using different evaluation strategies and found significant differences in terms of conformal performance measures. We also proposed several visualizations based on matrices, graphs, and charts that capture different aspects of the resulting prediction sets.}

\keywords{conformal-prediction, multi-user models, uncertainty estimation, visualization, machine-learning}



\maketitle

\section{Introduction}\label{sec:introduction}

With the advent of wearable devices and new kinds of sensors, it is now possible to monitor human behavior from different perspectives including physical activities~\citep{saleem2023}, hand gestures~\citep{rawat2023}, body postures~\citep{patel2017}, emotion~\citep{al2023}, sleep patterns~\citep{watson2021}, and even mental health such as stress and depression~\citep{garcia2018mental}. Most of those systems rely on machine learning models and are part of larger applications in human computer interaction, sports, behavioral research, and medicine, to name a few. Many of those applications play critical roles and thus, the machine learning algorithms supporting them need to be reliable and robust. The main thing in common about the aforementioned applications is that the data is intrinsic to the users. For example, in order for an emotion recognition system to be effective, it needs to be trained on real data from similar users from those that are expected to use the system. We call these types of scenarios where the data are inherent to the users: \emph{multi-user settings} or \emph{multi-user systems}. On the other hand, an object recognition system that identifies defects in machining parts or a weather forecast system are not multi-user systems.

Multi-user systems pose several challenges that are often overlooked being the two most important ones the \emph{inter/intra person variance}. That is, every person is different so we cannot expect a single general model to perform optimally (inter person variance). Moreover, a given person can also change their behavior over time and thus, a model that works now is not guaranteed to work in the future (intra person variance). 

In the literature, many works do not take into account the multi-user nature of a problem, as noted by~\cite{lockhart2014limitations}. That is, user identifier information is not used when deciding how to split the data when evaluating the generalization performance of a machine learning model. This typically results in overestimating the models' performance as will be explained in Section~\ref{sec:multi-user_validation}.

Another important aspect of a machine learning model that is intended to be deployed is its reliability. That is, \emph{To what extent can we trust the model's predictions?} Most machine learning models are designed to always produce an answer regardless of its validity. For example, in the traditional setting, a classifier will always produce a label. Even if the input data belongs to a new class for which there were no examples in the training set. Even worse, the predicted label could have an associated high confidence score which can mislead the final user resulting in potential negative consequences, specially in critical applications such as medicine or unmanned vehicles. Thus, it becomes of paramount importance the ability of a model to inform when it doesn't know the answer or when in doubt between different alternatives (labels for a classification problem) and this is exactly what \emph{conformal prediction} does. Conformal prediction is a general framework proposed by~\cite{vovk2005algorithmic} that extends the capabilities of any classifier or regression model such that the uncertainty of the predictions can be quantified. In the case of classification instead of predicting a single class, a conformal classifier predicts a set (possibly empty) of classes such that a performance measure is guaranteed to be satisfied. In the case of regression, the predictions are confidence intervals. More about the basics of conformal prediction will be covered in Section~\ref{sec:conformal_prediction}.

We believe that it should become a practice to evaluate interactive systems based in user data in terms of their 1) multi-user nature, and 2) their reliability. To this extent, in this work we make an evaluation of different applications in the intersection between their multi-user nature and predictions reliability within the conformal prediction framework. Based on our literature review, this is the first work that analyzes multi-user systems with conformal prediction. We also introduced a new evaluation strategy (user-calibrated model) which uses data from the target user to calibrate the conformal model. The main contributions of this work are threefold:

\begin{itemize}
    \item We conducted extensive experiments using four real-world datasets to investigate the influence of the choice of multi-user evaluation strategies (see Section~\ref{sec:multi-user_validation}) on the quality of conformal and traditional models in terms of conformal performance measures.
    \item We introduced a new evaluation strategy (user-calibrated model) which uses data from the target user to calibrate the conformal model. Based on our experiments, we found that a mixed-model (which is the most commonly used in the literature) overestimates the performance of a system compared to a user-independent model. The user-independent model is what you would usually experience in real life where the system does not have any information about the target user.
    \item We propose the use of matrix and graph based visualizations to analyze conformal prediction sets and compare the results between traditional and conformal models. Furthermore, in order to overcome the binary-relation limitations of the former visualizations, we propose the use of a multiset-enumerating chart for visualizing n-ary relations.
\end{itemize}

The specific null hypotheses we aim to test in this work with respect to \emph{coverage} are:

\begin{enumerate}
    \item There is no difference in performance between a user-independent model and a mixed model and user-dependent model. 
    \item There is no difference in performance between a user-dependent model and a mixed model and user-independent model.
    \item There is no difference between a user-calibrated model and a user-independent model.
\end{enumerate}

The aforementioned types of models are described in section~\ref{sec:multi-user_validation}. We conducted the experiments on four representative open datasets of multi-user settings and the code to reproduce the results is available here: \url{https://github.com/enriquegit/conformal-prediction-multiuser}.

This document is organized as follows: Section~\ref{sec:background} begins by presenting the background about multi-user validation strategies. Then, it continuous with an introduction to conformal prediction and performance measures (traditional and conformal). Section~\ref{sec:conformal_vis} describes the proposed visualization techniques to plot prediction sets results. In section~\ref{sec:datasets} we describe the datasets used in this study. Section~\ref{sec:experiments} presents the experiments and results. Finally, section~\ref{conslusions} presents the conclusions.

\section{Background}\label{sec:background}

This section presents the background concepts related to multi-user systems, conformal prediction, and performance measures for conformal prediction.

\subsection{Multi-User Validation Strategies}\label{sec:multi-user_validation}

As previously mentioned, traditional evaluation methods applied to an entire dataset (hold out, cross validation, etc.) do not provide good estimates in multi-user systems due to the variation between users. \cite{lockhart2014limitations} pointed out that from the papers they surveyed, $40\%$ did not make any distinction between users when evaluating their models. This evaluation strategy overestimates the generalization performance of a model since by randomly splitting the dataset into training and test sets, it is very likely that data from a given user will end up in both sets. This has the effect that the trained model already has some information about the given user and will perform relatively well. But when tested on a completely new user the performance will drop significantly~\citep{lockhart2014limitations}.

To this end, the distinction between several evaluation strategies should be taken into account when validating multi-user systems. The three main evaluation strategies based on the type of data partitioning are \emph{mixed}, \emph{user-independent}, and \emph{user-dependent} \citep{GarciaCejaBook} and typically, the resulting trained models are called \emph{mixed model}, \emph{user-independent model}, and \emph{user-dependent model}, respectively.

\begin{figure}[h!]
\centering
\includegraphics[width=0.6\textwidth]{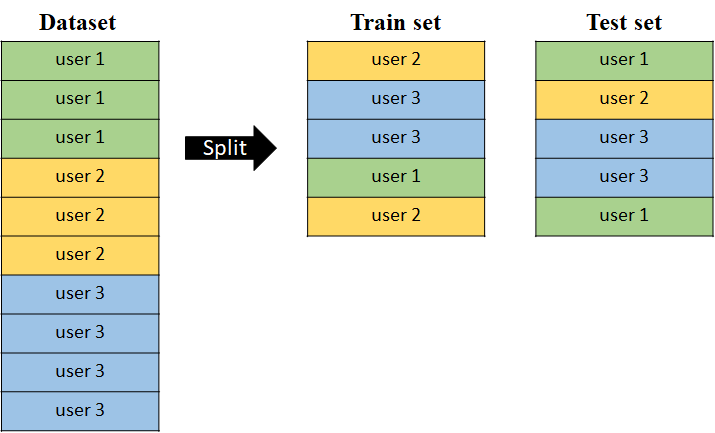}
\caption{Mixed Model data partitioning.}\label{fig:mixed_model}
\end{figure}

\textbf{Mixed Model (MM).} A mixed model is trained by partitioning the data irrespective of the users. That is, take all the data from all the users and randomly split it into training and test sets (Figure~\ref{fig:mixed_model}) or use cross validation. By doing so, it is very probable that part of the data from a given user $u$ will end up in the training set and some other in the test set. This means that at test time, the trained model will already `know' something about user $u$. Thus, we should expect the model to have a good performance since the data in the train and test sets are very similar. However when tested with a completely new user, the performance will drop.

\textbf{User-independent Model (UIM).} A user-independent Model doesn't know anything about the target user. That is, it is guaranteed that data from any potential user at test time was not present in the training set. This is the case when a system is to be used out-of-the-box without going into any calibration step specific to the final user. When the amount of data is limited, these models are evaluated using a leave-one-user-out strategy. That is, iterate trough each user $u$, and train the model with data from every user different from $u$ (Figure~\ref{fig:uim}). Compared to the mixed model and the user-dependent model (presented below), this one typically performs the worst.

\begin{figure}[h!]
\centering
\includegraphics[width=0.6\textwidth]{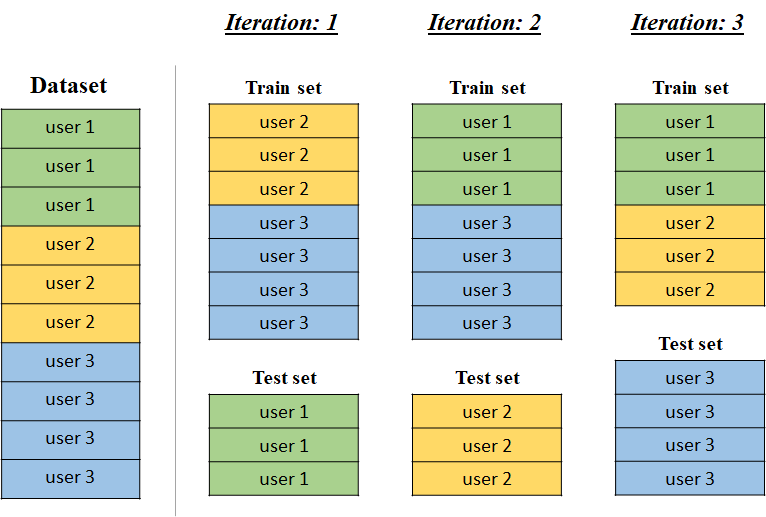}
\caption{User-independent Model data partitioning.}\label{fig:uim}
\end{figure}

\begin{figure}[h!]
\centering
\includegraphics[width=0.5\textwidth]{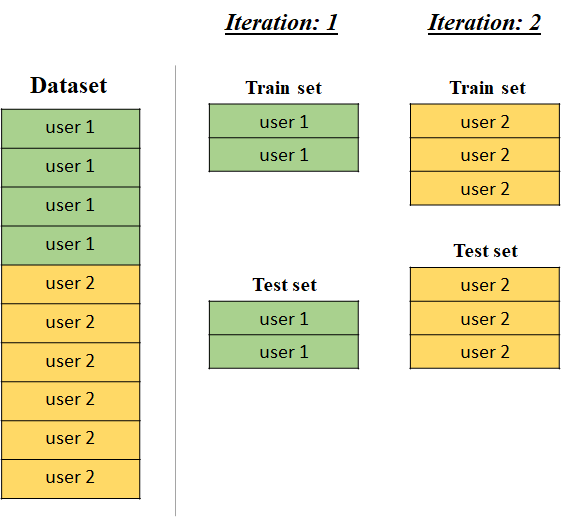}
\caption{User-dependent Model data partitioning.}\label{fig:udm}
\end{figure}

\textbf{User-dependent Model (UDM).} A user-dependent Model is trained with only data belonging strictly to the same user to be tested. This model typically performs the best since it is specialized for a particular user. To evaluate such models, iterate trough each user $u$ and train and test the model only with data from $u$. The data partitioning within each user can be done with hold-out or cross validation. Figure~\ref{fig:udm} shows this evaluation strategy using hold-out validation ($50$/$50$).

Apart from these, there is another type of model called \emph{user-adaptive model}. This one is a hybrid between the UIM and UDM. A user-adaptive model can start as a UIM and as more data from the target user is available, it is fined tuned to adapt it~\citep{churaev2022multi,garcia2020adaptive,siirtola2019incremental}. Another approach is to train the model using data only from the most similar users~\citep{xu2014cluster,ferrari2020personalization,garcia2015automatic} or to look for class similarities~\citep{garcia2015building}. In this work, user-adaptive models are not considered since they introduce some complexities namely: 1) most of them are model dependent, and 2) they are application dependent.

\subsection{Conformal Prediction}\label{sec:conformal_prediction}

In the last years, machine learning algorithms have permeated our everyday lives. With the recent advent of new hardware and software technologies, more complex and bigger predictive models are being developed capable of solving complex tasks and problems. However, despite this incredible progress, most machine learning models are designed to always produce an output regardless of its quality or trustworthiness. This imposes serious concerns in critical applications like in medical diagnosis, autonomous vehicles, fraud detection, to name a few. One of the aspects for a model to be robust, is its capability to accurately quantify the uncertainty of its predictions. However this aspect is typically not considered and rather, the focus is put on the overall performance of a model in terms of metrics such as precision, recall, F1-score, etc. While those metrics are useful for assessing the expected generalization performance of a model, they do not capture the expected average uncertainty of individual predictions (confidence intervals for regression and p-values for classification). In classification, most models can produce confidence scores for their predictions. For example, the output of the softmax function in a neural network~\citep{hendrycks2016baseline}. However, those confidence scores do not represent true probabilities, that is, they are uncalibrated~\citep{guo2017calibration,minderer2021revisiting}. Several calibration methods have been proposed \citep{karandikar2021soft,mukhoti2020calibrating} however they are specific for neural networks and often require modifications to the underlying model. On the other hand, there is a general method called \emph{conformal prediction} that was proposed by \cite{vovk2005algorithmic} that addresses those challenges.

Conformal prediction provides confidence estimates for individual predictions. Confidence intervals for regression and prediction sets for classification. As opposed to traditional classifiers that output a single prediction for every input data point, conformal prediction produces prediction sets. A prediction set contains the predicted labels such that the true label is contained with probability $1 - \epsilon$ where $\epsilon$ is the acceptable error level. The size of the prediction set reflects the confidence. Bigger set sizes reflect more uncertainty, thus, small sets are preferred. Conformal prediction works as a `wrapper' with any regression or classification model making it very flexible. Formally, a conformal predictor is defined as:

\begin{equation} \label{eq:distance_matrix}
 \Gamma^{\epsilon} \left( x_1,y_1,\ldots,x_{n-1},y_{n-1}, x_n \right) = y \in Y
\end{equation}

such that

\begin{equation} \label{eq:distance_matrix}
 \frac{\lvert \{i=1,\ldots,n-1:\alpha_i \geq \alpha_n \} \rvert}{n-1} > \epsilon
\end{equation}

where $y \in Y$ is the set of all confident labels and $\alpha_i$ is the nonconformity score for element $i$. A nonconformity score measures how different a data point is from previous examples. A typical nonconformity measure for classification is $1$ minus the confidence score of the predicted class. The quality and tightness of the predicted sets thus primarily depend on the expressiveness of the score for the given dataset and thus the choice of the score-producing algorithm. Bagging-based models such as Random Forests and Neural Network Ensembles as well as other probabilistic models generally produce more meaningful confidence scores and are thus well suited for conformal prediction ~\citep{jourdan_neurips_2021}.
The nonconformity scores $\alpha_i$ are estimated from a \emph{calibration set} which is independent from the training set.

The $\epsilon$ is defined by the user beforehand and determines what is the maximum tolerated error (or a confidence of $\gamma = 1-\epsilon$). Conformal models are guaranteed to produce predictions with a marginal error no more than $\epsilon$ given some assumptions, with exchangeability being one of the most critical in the context of multi-user applications.
Exchangeability is a statistical property of a sequence of random variables $( X_1, X_2, \ldots, X_n )$, where the joint probability distribution remains unchanged under any permutation $\pi$ of the sequence ~\cite{vovk2005algorithmic}:
\begin{equation}
P(X_1 = x_1, X_2 = x_2, \ldots, X_n = x_n) = P(X_{\pi(1)} = x_1, X_{\pi(2)} = x_2, \ldots, X_{\pi(n)} = x_n).
\end{equation}

This assumption is fundamental in generating reliable and statistically valid prediction intervals. In case of multi-user applications and MM as well as UIM models, \textit{cf.} Section \ref{sec:multi-user_validation},  in particular, this assumption may be violated by shifts in the data distributions of different users. This aspect is thus further analyzed in the experiment section of this work.

In addition to the fundamental assumption of exchangeability, several other critical assumptions underpin the conformal prediction framework. The nonconformity measure \( \alpha \), which is pivotal in assessing how well a new instance conforms to a set of existing data points, should exhibit a consistent behavior. Specifically, for any two instances \( x_i \) and \( x_j \), if they are similar under a chosen metric, their respective nonconformity scores \( \alpha_i \) and \( \alpha_j \) should be closely aligned. This ensures that the nonconformity measure reliably reflects the data distribution. The calibration of confidence levels, denoted as \( \gamma = 1 - \epsilon \), also plays a crucial role. It is imperative that these confidence levels are accurately calibrated so that the empirical coverage probability of the prediction intervals aligns with the intended confidence level. Practically, for a given confidence level \( \gamma \), the proportion of instances where the true value is contained within the interval should converge to \( \gamma \) with an increasing sample size. While not explicitly stated as an assumption, the effectiveness of conformal prediction is often contingent on the availability of a large sample size. A substantial dataset stabilizes the empirical distribution of nonconformity scores, thereby enhancing the robustness of the prediction intervals.

Conformal models are typically not solely evaluated by metrics that rely on the error level since this is defined by the user. Instead, they are evaluated based on properties of the prediction sets. In the next section, we describe some conformal prediction performance measures.

\subsection{Performance Measures}\label{sec:performance_metrics}

In this section we detail the performance measures that are used in this work. First, we describe standard non-conformal classification metrics based on the confusion matrix. Next, we discuss other measures that have been specifically designed to evaluate the quality of conformal predictors.

\subsubsection{Non-conformal Metrics}
In the assessment of predictive models for classification, several metrics are pivotal for understanding their performance. These include accuracy, sensitivity, specificity, precision, recall, and the F1 score. Each metric illuminates different facets of model performance and is essential for a comprehensive evaluation.

\subsubsection*{Accuracy}
Accuracy is the simplest and most intuitive performance metric. It is the ratio of correctly predicted observations (both true positives and true negatives) to the total number of observations. The formula for accuracy is given by:
\begin{equation}
    \text{Accuracy} = \frac{\text{True Positives (TP)} + \text{True Negatives (TN)}}{\text{Total Observations}}
\end{equation}

\subsubsection*{Sensitivity (True Positive Rate)}
Sensitivity, also known as the true positive rate, measures the proportion of actual positives that are correctly identified by the model. It is particularly useful for evaluating the effectiveness of a model in identifying positive outcomes. The formula for sensitivity is:
\begin{equation}
    \text{Sensitivity} = \frac{\text{TP}}{\text{TP} + \text{False Negatives (FN)}}
\end{equation}

\subsubsection*{Specificity (True Negative Rate)}
Specificity, or the true negative rate, measures the proportion of actual negatives that are correctly identified as such. This metric is crucial for assessing the model's ability to identify negative outcomes. The formula for specificity is:
\begin{equation}
    \text{Specificity} = \frac{\text{TN}}{\text{TN} + \text{False Positives (FP)}}
\end{equation}

\subsubsection*{Precision}
Precision is the ratio of correctly predicted positive observations to the total predicted positives. It is a measure of the accuracy of the positive predictions made by the model. The formula for precision is:
\begin{equation}
    \text{Precision} = \frac{\text{TP}}{\text{TP} + \text{FP}}
\end{equation}

\subsubsection*{Recall}
Recall, which is synonymous with sensitivity, measures the proportion of actual positives that are correctly identified. It focuses on the model's capability to capture positive cases. The formula for recall is the same as that for sensitivity:
\begin{equation}
    \text{Recall} = \frac{\text{TP}}{\text{TP} + \text{FN}}
\end{equation}

\subsubsection*{F1 Score}
The F1 score is the harmonic mean of precision and recall, providing a balance between these two metrics. It is particularly useful when the class distribution is imbalanced. The F1 score is calculated as follows:
\begin{equation}
    \text{F1 Score} = 2 \times \frac{\text{Precision} \times \text{Recall}}{\text{Precision} + \text{Recall}}
\end{equation}

\subsubsection{Conformal Measures}
While standard metrics like accuracy, sensitivity, and specificity provide a baseline understanding of model performance, evaluating conformal predictors require specific measures that reflect their unique characteristics. Conformal prediction, with its emphasis on reliable uncertainty quantification, necessitates measures that capture the quality of prediction intervals and the calibration of probabilistic outputs. This subsection delves into these specialized measures, explaining their significance and utility in assessing the performance of conformal prediction models. All these measures (except for the Jaccard index) are discussed in detail by \cite{vovk2016criteria}.



We start with a set of measures which consider only the composition of the prediction set, disregarding whether the expected class is part of the prediction set or not.

\subsection*{N criterion (set size)}
The \emph{N criterion} (\emph{setsize}) considers the size of the prediction sets, reporting the average thereof. Smaller values are preferred.
\begin{equation}
    \frac{1}{k} \sum_{i=1}^k |\Gamma^\epsilon_i|
\end{equation}

where $k$ is the number of prediction sets.

\subsubsection*{Percentage of Empty Sets}
The percentage of empty sets (\emph{pctempty}) records the proportion of empty prediction sets:
\begin{equation}
    \frac{1}{k} \sum_{i=1}^k \begin{cases}
      1, & \text{if}\ |\Gamma^\epsilon_i| = 0 \\
      0, & \text{otherwise}
    \end{cases}
\end{equation}

\subsubsection*{M criterion}
The M criterion is the proportion of prediction sets that contain more than one element:
\begin{equation}
    \frac{1}{k} \sum_{i=1}^k \begin{cases}
      1, & \text{if}\ |\Gamma^\epsilon_i| > 1 \\
      0, & \text{otherwise}
    \end{cases}
\end{equation}

\subsubsection*{F criterion}
This is the average fuzziness where the fuzziness for a test object $x_i$ refers to the sum of the p-values, excluding the largest one. The \emph{F criterion} reports the average fuzziness values for a sequence of prediction sets. Smaller values are considered to be better.

\begin{equation}
    \frac{1}{k} \sum_{i=1}^k \left( \sum_y p^y_i - \underset{y}{\mathrm{max}} p^y_i \right)
\end{equation}

\hfill\\

We now consider another set of measures, where the presence of the actual/true class is expected to be part of the prediction set. For this reason, the measures are also referred to as ``observed criteria''. In the following, we assume that $y_i$ corresponds with the actual/true class for pair $(x_i, y_i)$ in the test dataset.

\subsubsection*{Coverage}
Coverage refers to the percentage of prediction sets that contain the true class:
\begin{equation}
    \frac{1}{k} \sum_{i=1}^k \begin{cases}
      1, & \text{if}\ y_i \in \Gamma^\epsilon_i \\
      0, & \text{otherwise}
    \end{cases}
\end{equation}

\subsubsection*{Jaccard index}
The \emph{Jaccard index} is a commonly-used metric that measures the similarity of sets. A value of $1$ would indicate that a prediction is identical to the singleton set that includes the actual/true class. We consider computing the average value, and prefer a value close to $1$:
\begin{equation}
    \frac{1}{k} \sum_{i=1}^k \frac{|\Gamma^\epsilon_i \cap \{y_i\}|}{|\Gamma^\epsilon_i \cup \{y_i\}|}
\end{equation}

\subsection*{OM criterion}
The \emph{OM criterion} is the proportion of prediction sets that include one or more false classes. Smaller values are preferred.

\begin{equation}
    \frac{1}{k} \sum_{i=1}^k \begin{cases}
      1, & \text{if}\ \{\Gamma^\epsilon_i \diagdown \{y_i\} \neq \emptyset\} \\
      0, & \text{otherwise}
    \end{cases}
\end{equation}

\subsection*{OF criterion (observed fuzziness)}
The \emph{OF criterion} is the average of the sum of the p-values of the false classes in the prediction set:

\begin{equation}
    \frac{1}{k} \sum_{i=1}^k \sum_{y\neq y_i} p^y_i
\end{equation}

\subsection*{OU criterion (observed unconfidence)}
The \emph{OU criterion} is the average unconfidence. The unconfidence of a prediction set is the largest p-value of the false classes.

\begin{equation}
    \frac{1}{k} \sum_{i=1}^k \underset{y\neq y_i}{\mathrm{max}}\ p^y_i
\end{equation}

\subsection*{OE criterion (observed excess)}
The \emph{OE criterion} is the average number of false classes in the prediction sets.

\begin{equation}
    \frac{1}{k} \sum_{i=1}^k |\Gamma^\epsilon_i \diagdown \{y_i\}|
\end{equation}

\section{Prediction Sets Visualizations}\label{sec:conformal_vis}

In this section we describe the methods we will use to visualize the results of the conformal prediction sets. Non-conformal single predictions are typically visualized with a \emph{confusion matrix}. Here, we assume that the rows represent the predicted classes and columns represent the true classes. Specifically, we propose four visualizations: 1) co-occurrence matrix, 2) co-occurrence graph, 3) zero diagonal confusion matrix (ZDCM), and 4) multiset-enumerating chart. The first two are intended to visualize prediction sets while the ZDCM is a modification of the traditional confusion matrix that will allow us to compare the relationship between non-conformal and conformal predictions. The fourth one refers to a family of charts which enumerates all, or part of all, possible prediction sets along with a chart which displays statistics about the underlying predictions.

We will also make use of the traditional confusion matrix. Whenever we refer to a confusion matrix, we will assume that it is column-normalized. That is, every column sums up to $1$. See Figure~\ref{fig:d1-plots-rf-mm} (middle-right).

\textbf{Zero diagonal confusion matrix (ZDCM).} This is a traditional confusion matrix but with the diagonal set to $0$ before normalization. That is, its columns still sum up to $1$. Note that it is not symmetric. The rationale behind the ZDCM is to highlight which classes were confused with which other ones. Figure~\ref{fig:d1-plots-rf-mm} (middle-left) shows a ZDCM. Here we can see that the class \emph{`Sitting'} was confused $50\%$ of the time with \emph{`Standing'} and $50\%$ with \emph{`Downstairs'}. Later, we will see how the ZDCM allows us to compare single predictions with prediction sets.

\textbf{Co-occurrence matrix.} This visualization is intended to highlight which pairs of classes occurred together in the prediction sets. This is a $n\times n$ matrix where $n$ is the number of classes. Every column represents a class and the proportion of times that class appeared together with every other class (excluding itself). The proportions that a class appears with itself are excluded because otherwise, the diagonal would dominate and the off-diagonal patterns would fade away making them difficult to visualize. The columns of this matrix also sum up to $1$ and the diagonal is $0$. This matrix is not symmetric. Figure~\ref{fig:d1-plots-rf-mm} (top-left) shows an example of a co-occurrence matrix. For example, we can see that \emph{`Jogging'} occurred together with \emph{`Downstairs'} $16.1\%$, with \emph{`Upstairs'} $68.9\%$, and with \emph{`Walking'} $15\%$. It can also be noted that the co-occurrence matrix is related to the ZDCM. That is, pairs of classes that tend to be confused, are also likely to appear in the same prediction set.

\textbf{Co-occurrence graph.} This visualization is an undirected graph where nodes represent classes and edges represent counts of how many times a pair of classes occurred together in the prediction sets. The thickness of the edge is proportional to the counts and the size of a node is proportional to its degree. This visualization allow us to understand which classes frequently appeared along other classes (`popular classes'). Figure~\ref{fig:d1-plots-rf-mm} (top-right) shows a co-occurrence graph. This graph can be compared with a confusion matrix. For instance, popular classes (bigger nodes) also tend to be more colored in the confusion matrix (rows).

\textbf{Multiset-enumerating chart.}
Given that the output of conformal prediction takes the form of sets, a visualization based on binary relations might obfuscate facts and even lead to wrong conclusions. For instance, the thickness of the edge connecting \emph{`Downstairs'} and \emph{`Upstairs'} in the co-occurrence graph results from the frequency of the prediction set $\left\{ Downstairs, Upstairs\right\}$ added to the frequency of $\left\{ Downstairs, Upstairs, Walking\right\}$. There is no way to distinguish the source of the thickness. For this reason, we introduce the multiset-enumerating chart. This chart, which is an adaptation of the UpSet chart \citep{2014_infovis_upset}, enumerates explicitly the powerset of the set of classes, and associates each set with a chart/plot. Figure \ref{fig:d1-plots-rf-mm} (bottom-center) presents an example of a multiset-enumerating chart which displays the frequency of each prediction set for a given dataset. From that example, we can draw some clear conclusions: 1) the prediction sets were dominated by singletons, 2) the sets $\left\{Sit, Stand\right\}$ and $\left\{Stand, Walk\right\}$ were also frequently predicted, implying that \emph{`Stand'} can be confounded with either \emph{`Walk'}, 3) the rest of sets are rarely predicted. One shortcoming of this visualization is the size of the powerset can be large, i.e., $2^{|C|}$, where $C$ is the set of classes. This inconvenient can be alleviated by filtering out low frequency sets.

\section{Datasets}\label{sec:datasets}

In this section we describe each of the datasets used for the evaluation. We used the following criteria for including a given dataset:

\begin{itemize}
    \item The dataset was collected by at least two users.
    \item The dataset contains information such that it is possible to determine which sensor readings were collected by which specific user.
    \item At least $5$ instances per user-class pair exist.
\end{itemize}

In all datasets, we only kept users that have at least $5$ repetitions for every class.

\subsection{D1: WISDM}\label{d1}

The WISDM dataset was collected using Android based smartphones \citep{Kwapisz2011,wisdm}. The data captured with the embedded tri-axial accelerometer was recorded while the users performed $6$ different physical activities (Walking, Jogging, Walking downstairs, Walking upstairs, Sitting, and Standing). The phone was placed in the users' pocket and the sampling rate was set to $20$Hz. They already provided a version with $46$ extracted features from fixed length windows of $10$ seconds each. Some of the features are the average (for each axis), average resultant acceleration, time between peaks, binned distribution, standard deviation (for each axis), etc. Some columns ($5$) were removed due to corrupted data and missing values. The dataset originally has $36$ users. After filtering users to meet the criteria, $16$ users were kept and the total number of instances were $2,869$. Table~\ref{tab:dist_wisdm} shows the final class distribution.

\begin{table}[h!]
\centering
\caption{Class distribution for D1: WISDM.}
\begin{tabular}{ccc}
\toprule
\textbf{Classes} & \textbf{Number of instances} & \textbf{Proportion of instances} \\
\midrule
\textbf{Downstairs} & \multicolumn{1}{c}{300} & \multicolumn{1}{c}{10.45\%} \\
\textbf{Jogging} & \multicolumn{1}{c}{833} & \multicolumn{1}{c}{29.0\%} \\
\textbf{Sitting} & \multicolumn{1}{c}{241} & \multicolumn{1}{c}{8.4\%} \\
\textbf{Standing} & \multicolumn{1}{c}{176} & \multicolumn{1}{c}{6.13\%} \\
\textbf{Upstairs} & \multicolumn{1}{c}{329} & \multicolumn{1}{c}{11.47\%} \\
\textbf{Walking} & \multicolumn{1}{c}{990} & \multicolumn{1}{c}{34.5\%} \\
\bottomrule
\end{tabular}
\label{tab:dist_wisdm}
\end{table}

\subsection{D2: HAR70+}\label{d2}

The HAR70+ dataset was collected using two tri-axial accelerometers by $18$ older adult participants ($70$-$95$ years old) \citep{har70,Ustad2023}. One of the sensors was attached to the right front thigh while the other one was in the lower back. The sampling rate was set to $50$Hz. The dataset has $7$ activities: Walking, Shuffling, Ascending stairs, Descending stairs, Standing, Sitting, and Lying. Since data for activities Ascending/Descending stairs was scarce, those classes were removed. For each sensor, $16$ features were extracted from non-overlapping windows of size $150$ ($3$ seconds). The features included: mean, standard deviation, and maximum for each axis. The correlation between each pair of axes, the mean magnitude, the standard deviation of the magnitude, area under the curve, and the mean difference between consecutive readings. A moving average filter of size $10$ was applied before extracting the features. After filtering the users that don't meet the criteria, $16$ users were kept in the dataset and a total of $12,887$ instances were generated. Table~\ref{tab:dist_har70} shows the final class distribution.

\begin{table}[h!]
\centering
\caption{Class distribution for D2: HAR70+.}
\begin{tabular}{ccc}
\toprule
\textbf{Classes} & \textbf{Number of instances} & \textbf{Proportion of instances} \\
\midrule
\textbf{Lying} & \multicolumn{1}{c}{1096} & \multicolumn{1}{c}{8.5\%} \\
\textbf{Shuffling} & \multicolumn{1}{c}{405} & \multicolumn{1}{c}{3.14\%} \\
\textbf{Sitting} & \multicolumn{1}{c}{2456} & \multicolumn{1}{c}{19.0\%} \\
\textbf{Standing} & \multicolumn{1}{c}{2492} & \multicolumn{1}{c}{19.34\%} \\
\textbf{Walking} & \multicolumn{1}{c}{6438} & \multicolumn{1}{c}{50\%} \\
\bottomrule
\end{tabular}
\label{tab:dist_har70}
\end{table}

\subsection{D3: Opportunity}\label{d3}

This dataset consists of daily activities recorded with body-worn sensors, object sensors and ambient sensors \citep{Roggen2010,opp}. In this work we use the locomotion actions labels (Stand, Walk, Sit, Lie). The data was recorded by $4$ users. Even though the users worn several sensors, we only extracted features from two accelerometers as in \citep{garcia2018improved} which were sampled at $30$Hz. One located in the back and the other one in the right shoe. The same $16$ features as the D2: HAR70+ dataset were extracted for each sensor on fixed non-overlapping windows of size 120 (4 seconds). The total number of resulting instances were $4,080$. Table~\ref{tab:dist_opportunity} shows the final class distribution.

\begin{table}[h!]
\centering
\caption{Class distribution for D3: Opportunity.}
\begin{tabular}{ccc}
\toprule
\textbf{Classes} & \textbf{Number of instances} & \textbf{Proportion of instances} \\
\midrule
\textbf{Lying} & \multicolumn{1}{c}{202} & \multicolumn{1}{c}{5.0\%} \\
\textbf{Sitting} & \multicolumn{1}{c}{977} & \multicolumn{1}{c}{24.0\%} \\
\textbf{Standing} & \multicolumn{1}{c}{1832} & \multicolumn{1}{c}{45.0\%} \\
\textbf{Walking} & \multicolumn{1}{c}{1069} & \multicolumn{1}{c}{26.0\%} \\
\bottomrule
\end{tabular}
\label{tab:dist_opportunity}
\end{table}

\subsection{D4: Smartwatch Gestures}\label{d4}

The smartwatch gestures dataset was collected to evaluate several gesture recognition algorithms for mobile application control using arm gestures. The dataset was originally presented in \citep{smartwatch_gestures} and contains acceleration data from the $3$-axis accelerometer of a first generation Sony SmartWatch at a sampling rate of $10$ Hz paired with measurement timestamps. The device was worn on the right wrist of the users. In total, the dataset covers $20$ different gestures, performed $20$ times by $8$ different users. The class frequencies are thus approximately uniformly distributed. The gestures have been manually segmented by the SmartWatch users. The same $16$ features as in the previous two datasets were extracted for each of the gesture samples. Table~\ref{tab:smartwatch_gestures} shows the final class distribution.

\begin{table}[h!]
\centering
\caption{Class distribution for D4: Smartwatch Gestures.}
\begin{tabular}{ccc}
\toprule
\textbf{Classes} & \textbf{Number of instances} & \textbf{Proportion of instances} \\
\midrule
\textbf{Gesture 01} & \multicolumn{1}{c}{160} & \multicolumn{1}{c}{4.9\%} \\
\textbf{Gesture 02} & \multicolumn{1}{c}{161} & \multicolumn{1}{c}{5.0\%} \\
\textbf{Gesture 03} & \multicolumn{1}{c}{166} & \multicolumn{1}{c}{5.1\%} \\
\textbf{Gesture 04} & \multicolumn{1}{c}{166} & \multicolumn{1}{c}{5.1\%} \\
\textbf{Gesture 05} & \multicolumn{1}{c}{160} & \multicolumn{1}{c}{4.9\%} \\
\textbf{Gesture 06} & \multicolumn{1}{c}{164} & \multicolumn{1}{c}{5.0\%} \\
\textbf{Gesture 07} & \multicolumn{1}{c}{161} & \multicolumn{1}{c}{5.0\%} \\
\textbf{Gesture 08} & \multicolumn{1}{c}{164} & \multicolumn{1}{c}{5.0\%} \\
\textbf{Gesture 09} & \multicolumn{1}{c}{161} & \multicolumn{1}{c}{5.0\%} \\
\textbf{Gesture 10} & \multicolumn{1}{c}{164} & \multicolumn{1}{c}{5.0\%} \\
\textbf{Gesture 11} & \multicolumn{1}{c}{162} & \multicolumn{1}{c}{5.0\%} \\
\textbf{Gesture 12} & \multicolumn{1}{c}{161} & \multicolumn{1}{c}{5.0\%} \\
\textbf{Gesture 13} & \multicolumn{1}{c}{161} & \multicolumn{1}{c}{5.0\%} \\
\textbf{Gesture 14} & \multicolumn{1}{c}{168} & \multicolumn{1}{c}{5.2\%} \\
\textbf{Gesture 15} & \multicolumn{1}{c}{162} & \multicolumn{1}{c}{5.0\%} \\
\textbf{Gesture 16} & \multicolumn{1}{c}{161} & \multicolumn{1}{c}{5.0\%} \\
\textbf{Gesture 17} & \multicolumn{1}{c}{163} & \multicolumn{1}{c}{5.0\%} \\
\textbf{Gesture 18} & \multicolumn{1}{c}{163} & \multicolumn{1}{c}{5.0\%} \\
\textbf{Gesture 19} & \multicolumn{1}{c}{162} & \multicolumn{1}{c}{5.0\%} \\
\textbf{Gesture 20} & \multicolumn{1}{c}{161} & \multicolumn{1}{c}{5.0\%} \\
\bottomrule
\end{tabular}
\label{tab:smartwatch_gestures}
\end{table}

\subsection{Summary of Datasets}\label{summary}

Table~\ref{tab:datasets_summary} presents a summary of all the datasets used in this study after filtering them using the criteria specified in Section~\ref{sec:datasets}.

\begin{table}[h!]
\small 
\centering
\caption{Datasets Summary}
\begin{tabular}{C{1cm}C{2cm}C{2cm}C{2.2cm}C{2cm}}
\toprule
\textbf{Abbr.} & \textbf{Name} & \textbf{\# subjects} & \textbf{\# classes} & \textbf{\# instances} \\
\midrule D1 & WISDM & 16 & 6 & $2,869$\\ 
\midrule D2 & HAR70+ & 16 & 5 & $12,887$\\ 
\midrule D3 & Opportunity & 4 & 4 & $4,080$\\ 
\midrule D4 & Smartwatch Gestures & 8 & 20 & $3,200$\\
\bottomrule
\end{tabular}
\label{tab:datasets_summary}
\end{table}

\section{Experiments and Results}\label{sec:experiments}

For the experiments, we evaluated four different settings: Mixed models, user-dependent models, user-independent models, and a user-calibrated models (UCM). The UCM is trained the same way as the User-independent model but it is calibrated only with data from the target user. The conformal confidence was set to $\alpha=0.95$.

For all datasets, the data was split in the same way based on the type of model. For the MM and UDM the training set consisted of $60\%$, test set $20\%$, and calibration set $20\%$. For the UIM, the training set consisted of $60\%$ and calibration set $40\%$ all these data excluding the target user. The test set consisted of $50\%$ of all of the target user data. For the UCM the training set was $60\%$ of all the data excluding the target user. The calibration set was $50\%$ of the target user data and the other $50\%$ of the target user data was assigned to the test set.
In order to account for variance, each experiment was run $20$ times. Since some of the classifiers are scale sensitive, the data was normalized to $0-1$. The scaling parameters were learned only from the training set in order to avoid data injection. The experiments were implemented with the help of the conformal prediction library MAPIE~\citep{mapie}. We used the LAC method as the non-conformity function defined as one minus the score of the true label\footnote{MAPIE documentation: \url{https://mapie.readthedocs.io/en/latest/theoretical_description_classification.html}}. We trained four types of conformal classifiers: Naive Bayes, Random Forest (RF), Support Vector Machine (SVM), and \emph{k}-nearest neighbors (KNN). 

Tables~\ref{tab:d1-mixed},\ref{tab:d1-UDM},\ref{tab:d1-UIM}, and \ref{tab:d1-UCM} show the average and standard deviation results across the $20$ experiments iterations for the MM, UDM, UIM, and UCM, respectively. The first four are non-conformal measures and are in percentages. The remaining measures are the conformal ones. For the non-conformal measures, RF consistently achieved the best results for all types of models. For all types of models, except the UIM the minimum coverage of $95\%$ was achieved. In conformal prediction, the calibration set is assumed to have the same distribution as the test set. For the UIM this assumption is violated since the model does not have any information about the target user. On the other hand, the minimum coverage for the UCM was attained for all classifiers. Even though the training set doesn't have data from the target user, only data from the target user was used for calibration. In this case, having calibration data from the given target user allowed the model to satisfy the minimum coverage. It is also interesting to see that the calibration set from a single user in the UCM is much smaller than the calibration set taken from all users in the UIM. Still with much less data, the coverage was satisfied.

In most of the cases (except in the UCM) RF achieved the smallest average set size. The biggest set sizes were produced by KNN (latter we will discuss why). The percent of empty sets was very low and in many cases, no empty sets were produced by the classifiers.

Also note that the non-conformal measures between the UIM and UCM are the same since the models were trained with the same data. However, their conformal measures differ because the calibration set was different.

\begin{table}[h!]
\caption{D1: Results MM. * Smaller values are preferred.}
\centering
\begin{tabular}[t]{lllll}
\toprule
  & Naive Bayes & RF & SVM & KNN\\
\midrule
\cellcolor{gray!6}{accuracy} & \cellcolor{gray!6}{78.65$\pm$1.10} & \cellcolor{gray!6}{88.92$\pm$0.91} & \cellcolor{gray!6}{79.89$\pm$0.73} & \cellcolor{gray!6}{88.28$\pm$1.18}\\
sensitivity & 70.02$\pm$1.68 & 84.20$\pm$1.34 & 69.97$\pm$1.29 & 82.88$\pm$1.87\\
\cellcolor{gray!6}{specificity} & \cellcolor{gray!6}{95.25$\pm$0.23} & \cellcolor{gray!6}{97.60$\pm$0.19} & \cellcolor{gray!6}{95.23$\pm$0.19} & \cellcolor{gray!6}{97.57$\pm$0.24}\\
F1 & 69.53$\pm$1.68 & 85.25$\pm$1.25 & 69.84$\pm$1.64 & 83.63$\pm$1.73\\
\cellcolor{gray!6}{coverage} & \cellcolor{gray!6}{95.46$\pm$1.26} & \cellcolor{gray!6}{95.43$\pm$1.28} & \cellcolor{gray!6}{95.22$\pm$0.93} & \cellcolor{gray!6}{98.22$\pm$2.28}\\
jaccard & 0.60$\pm$0.03 & 0.86$\pm$0.01 & 0.73$\pm$0.01 & 0.44$\pm$0.34\\
\cellcolor{gray!6}{setsize*} & \cellcolor{gray!6}{2.06$\pm$0.14} & \cellcolor{gray!6}{1.22$\pm$0.05} & \cellcolor{gray!6}{1.63$\pm$0.06} & \cellcolor{gray!6}{4.10$\pm$2.39}\\
pctempty & 0.00$\pm$0.00 & 0.10$\pm$0.13 & 0.00$\pm$0.00 & 0.00$\pm$0.00\\
\cellcolor{gray!6}{MCriterion*} & \cellcolor{gray!6}{0.58$\pm$0.04} & \cellcolor{gray!6}{0.20$\pm$0.04} & \cellcolor{gray!6}{0.40$\pm$0.03} & \cellcolor{gray!6}{0.69$\pm$0.39}\\
FCriterion* & 0.18$\pm$0.02 & 0.04$\pm$0.01 & 0.09$\pm$0.01 & 0.28$\pm$0.05\\
\cellcolor{gray!6}{OM*} & \cellcolor{gray!6}{0.60$\pm$0.03} & \cellcolor{gray!6}{0.23$\pm$0.03} & \cellcolor{gray!6}{0.42$\pm$0.03} & \cellcolor{gray!6}{0.70$\pm$0.38}\\
OF* & 0.17$\pm$0.02 & 0.03$\pm$0.01 & 0.10$\pm$0.01 & 0.24$\pm$0.15\\
\cellcolor{gray!6}{OU*} & \cellcolor{gray!6}{0.13$\pm$0.01} & \cellcolor{gray!6}{0.03$\pm$0.00} & \cellcolor{gray!6}{0.08$\pm$0.00} & \cellcolor{gray!6}{0.09$\pm$0.03}\\
OE* & 1.10$\pm$0.13 & 0.26$\pm$0.04 & 0.68$\pm$0.05 & 3.12$\pm$2.37\\
\bottomrule
\end{tabular}
\label{tab:d1-mixed}
\end{table}

\begin{table}[h!]
\caption{D1: Results UDM. * Smaller values are preferred.}
\centering
\begin{tabular}[t]{lllll}
\toprule
  & Naive Bayes & RF & SVM & KNN\\
\midrule
\cellcolor{gray!6}{accuracy} & \cellcolor{gray!6}{87.97$\pm$1.18} & \cellcolor{gray!6}{92.67$\pm$0.96} & \cellcolor{gray!6}{82.95$\pm$1.22} & \cellcolor{gray!6}{85.51$\pm$1.08}\\
sensitivity & 80.82$\pm$2.08 & 89.02$\pm$1.33 & 72.87$\pm$1.73 & 75.85$\pm$2.00\\
\cellcolor{gray!6}{specificity} & \cellcolor{gray!6}{97.53$\pm$0.26} & \cellcolor{gray!6}{98.43$\pm$0.21} & \cellcolor{gray!6}{95.88$\pm$0.27} & \cellcolor{gray!6}{96.95$\pm$0.23}\\
F1 & 85.56$\pm$2.25 & 90.44$\pm$1.38 & 86.69$\pm$1.66 & 82.89$\pm$1.30\\
\cellcolor{gray!6}{coverage} & \cellcolor{gray!6}{98.61$\pm$0.64} & \cellcolor{gray!6}{97.63$\pm$0.56} & \cellcolor{gray!6}{97.61$\pm$0.75} & \cellcolor{gray!6}{99.00$\pm$0.51}\\
jaccard & 0.33$\pm$0.07 & 0.79$\pm$0.06 & 0.68$\pm$0.03 & 0.33$\pm$0.06\\
\cellcolor{gray!6}{setsize*} & \cellcolor{gray!6}{4.77$\pm$0.52} & \cellcolor{gray!6}{1.71$\pm$0.36} & \cellcolor{gray!6}{2.04$\pm$0.21} & \cellcolor{gray!6}{4.90$\pm$0.44}\\
pctempty & 0.03$\pm$0.07 & 0.42$\pm$0.30 & 0.29$\pm$0.37 & 0.06$\pm$0.12\\
\cellcolor{gray!6}{MCriterion*} & \cellcolor{gray!6}{0.81$\pm$0.08} & \cellcolor{gray!6}{0.30$\pm$0.07} & \cellcolor{gray!6}{0.47$\pm$0.04} & \cellcolor{gray!6}{0.82$\pm$0.07}\\
FCriterion* & 0.37$\pm$0.05 & 0.17$\pm$0.01 & 0.20$\pm$0.01 & 0.46$\pm$0.03\\
\cellcolor{gray!6}{OM*} & \cellcolor{gray!6}{0.82$\pm$0.08} & \cellcolor{gray!6}{0.31$\pm$0.07} & \cellcolor{gray!6}{0.48$\pm$0.04} & \cellcolor{gray!6}{0.82$\pm$0.07}\\
OF* & 0.37$\pm$0.06 & 0.06$\pm$0.02 & 0.11$\pm$0.01 & 0.46$\pm$0.04\\
\cellcolor{gray!6}{OU*} & \cellcolor{gray!6}{0.12$\pm$0.02} & \cellcolor{gray!6}{0.03$\pm$0.01} & \cellcolor{gray!6}{0.07$\pm$0.01} & \cellcolor{gray!6}{0.14$\pm$0.01}\\
OE* & 3.78$\pm$0.51 & 0.73$\pm$0.35 & 1.06$\pm$0.21 & 3.91$\pm$0.44\\
\bottomrule
\end{tabular}
\label{tab:d1-UDM}
\end{table}

\begin{table}[h!]
\caption{D1: Results UIM. * Smaller values are preferred.}
\centering
\begin{tabular}[t]{lllll}
\toprule
  & Naive Bayes & RF & SVM & KNN\\
\midrule
\cellcolor{gray!6}{accuracy} & \cellcolor{gray!6}{75.45$\pm$0.85} & \cellcolor{gray!6}{78.03$\pm$0.96} & \cellcolor{gray!6}{75.58$\pm$0.83} & \cellcolor{gray!6}{69.13$\pm$0.84}\\
sensitivity & 66.98$\pm$0.92 & 72.74$\pm$1.00 & 66.27$\pm$0.87 & 62.58$\pm$0.98\\
\cellcolor{gray!6}{specificity} & \cellcolor{gray!6}{94.55$\pm$0.19} & \cellcolor{gray!6}{95.19$\pm$0.19} & \cellcolor{gray!6}{94.26$\pm$0.19} & \cellcolor{gray!6}{93.60$\pm$0.17}\\
F1 & 73.90$\pm$1.89 & 77.09$\pm$1.39 & 77.65$\pm$1.34 & 66.05$\pm$1.54\\
\cellcolor{gray!6}{coverage} & \cellcolor{gray!6}{93.89$\pm$0.50} & \cellcolor{gray!6}{88.76$\pm$1.35} & \cellcolor{gray!6}{91.83$\pm$0.49} & \cellcolor{gray!6}{88.04$\pm$2.70}\\
jaccard & 0.58$\pm$0.01 & 0.73$\pm$0.01 & 0.68$\pm$0.01 & 0.50$\pm$0.09\\
\cellcolor{gray!6}{setsize*} & \cellcolor{gray!6}{2.07$\pm$0.04} & \cellcolor{gray!6}{1.39$\pm$0.04} & \cellcolor{gray!6}{1.70$\pm$0.04} & \cellcolor{gray!6}{2.97$\pm$0.75}\\
pctempty & 0.00$\pm$0.00 & 0.09$\pm$0.11 & 0.00$\pm$0.00 & 0.00$\pm$0.00\\
\cellcolor{gray!6}{MCriterion*} & \cellcolor{gray!6}{0.58$\pm$0.02} & \cellcolor{gray!6}{0.34$\pm$0.03} & \cellcolor{gray!6}{0.43$\pm$0.02} & \cellcolor{gray!6}{0.58$\pm$0.11}\\
FCriterion* & 0.18$\pm$0.01 & 0.06$\pm$0.00 & 0.10$\pm$0.00 & 0.27$\pm$0.01\\
\cellcolor{gray!6}{OM*} & \cellcolor{gray!6}{0.62$\pm$0.01} & \cellcolor{gray!6}{0.42$\pm$0.02} & \cellcolor{gray!6}{0.48$\pm$0.02} & \cellcolor{gray!6}{0.65$\pm$0.09}\\
OF* & 0.19$\pm$0.01 & 0.07$\pm$0.00 & 0.13$\pm$0.00 & 0.27$\pm$0.04\\
\cellcolor{gray!6}{OU*} & \cellcolor{gray!6}{0.14$\pm$0.00} & \cellcolor{gray!6}{0.06$\pm$0.00} & \cellcolor{gray!6}{0.10$\pm$0.00} & \cellcolor{gray!6}{0.18$\pm$0.01}\\
OE* & 1.13$\pm$0.04 & 0.50$\pm$0.03 & 0.78$\pm$0.03 & 2.09$\pm$0.73\\
\bottomrule
\end{tabular}
\label{tab:d1-UIM}
\end{table}

\begin{table}[h!]
\caption{D1: Results UCM. * Smaller values are preferred.}
\centering
\begin{tabular}[t]{lllll}
\toprule
  & Naive Bayes & RF & SVM & KNN\\
\midrule
\cellcolor{gray!6}{accuracy} & \cellcolor{gray!6}{75.45$\pm$0.85} & \cellcolor{gray!6}{78.03$\pm$0.96} & \cellcolor{gray!6}{75.58$\pm$0.83} & \cellcolor{gray!6}{69.13$\pm$0.84}\\
sensitivity & 66.98$\pm$0.92 & 72.74$\pm$1.00 & 66.27$\pm$0.87 & 62.58$\pm$0.98\\
\cellcolor{gray!6}{specificity} & \cellcolor{gray!6}{94.55$\pm$0.19} & \cellcolor{gray!6}{95.19$\pm$0.19} & \cellcolor{gray!6}{94.26$\pm$0.19} & \cellcolor{gray!6}{93.60$\pm$0.17}\\
F1 & 73.90$\pm$1.89 & 77.09$\pm$1.39 & 77.65$\pm$1.34 & 66.05$\pm$1.54\\
\cellcolor{gray!6}{coverage} & \cellcolor{gray!6}{95.62$\pm$0.68} & \cellcolor{gray!6}{96.28$\pm$0.62} & \cellcolor{gray!6}{95.46$\pm$0.70} & \cellcolor{gray!6}{99.92$\pm$0.25}\\
jaccard & 0.58$\pm$0.02 & 0.61$\pm$0.01 & 0.65$\pm$0.01 & 0.17$\pm$0.01\\
\cellcolor{gray!6}{setsize*} & \cellcolor{gray!6}{2.27$\pm$0.15} & \cellcolor{gray!6}{2.04$\pm$0.09} & \cellcolor{gray!6}{1.95$\pm$0.08} & \cellcolor{gray!6}{5.97$\pm$0.09}\\
pctempty & 0.00$\pm$0.00 & 0.04$\pm$0.05 & 0.08$\pm$0.13 & 0.00$\pm$0.00\\
\cellcolor{gray!6}{MCriterion*} & \cellcolor{gray!6}{0.60$\pm$0.03} & \cellcolor{gray!6}{0.60$\pm$0.02} & \cellcolor{gray!6}{0.51$\pm$0.02} & \cellcolor{gray!6}{1.00$\pm$0.01}\\
FCriterion* & 0.25$\pm$0.01 & 0.21$\pm$0.01 & 0.20$\pm$0.01 & 1.00$\pm$0.04\\
\cellcolor{gray!6}{OM*} & \cellcolor{gray!6}{0.62$\pm$0.03} & \cellcolor{gray!6}{0.62$\pm$0.02} & \cellcolor{gray!6}{0.52$\pm$0.02} & \cellcolor{gray!6}{1.00$\pm$0.01}\\
OF* & 0.24$\pm$0.01 & 0.20$\pm$0.01 & 0.20$\pm$0.01 & 1.11$\pm$0.03\\
\cellcolor{gray!6}{OU*} & \cellcolor{gray!6}{0.16$\pm$0.01} & \cellcolor{gray!6}{0.15$\pm$0.01} & \cellcolor{gray!6}{0.13$\pm$0.00} & \cellcolor{gray!6}{0.35$\pm$0.01}\\
OE* & 1.32$\pm$0.15 & 1.08$\pm$0.09 & 0.99$\pm$0.08 & 4.97$\pm$0.08\\
\bottomrule
\end{tabular}
\label{tab:d1-UCM}
\end{table}

\begin{figure}[h!]
\centering
\includegraphics[width=.40\textwidth]{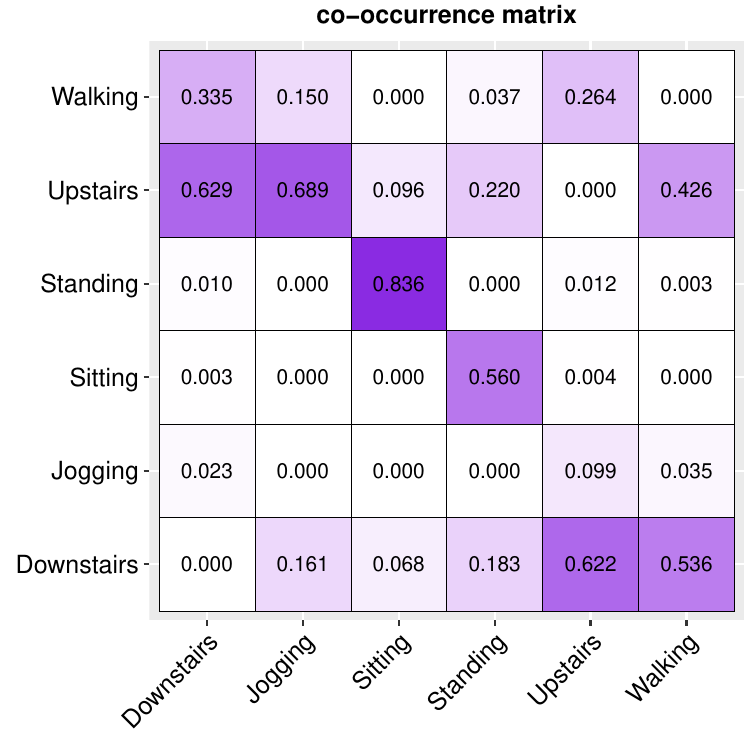}\hfill
\includegraphics[width=.40\textwidth]{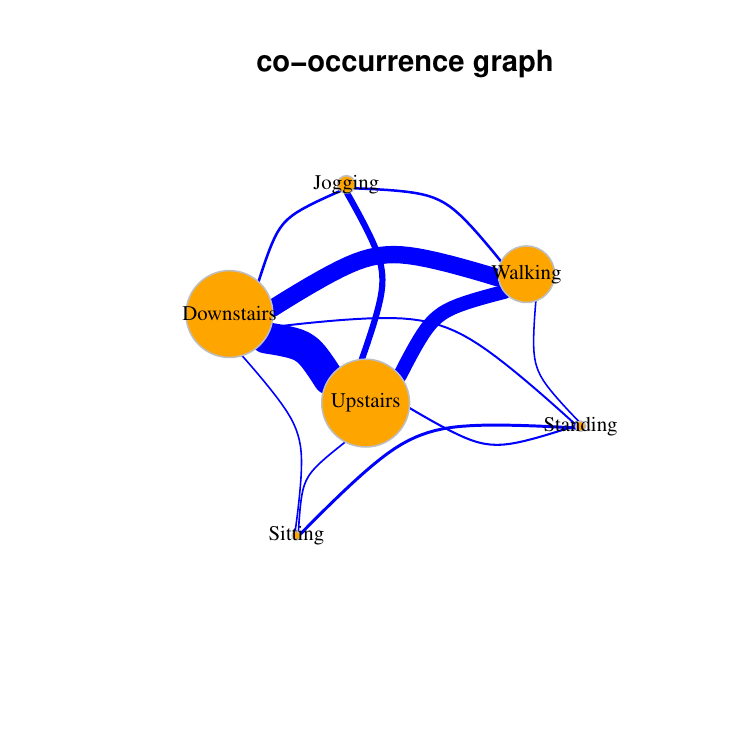}\hfill

\includegraphics[width=.40\textwidth]{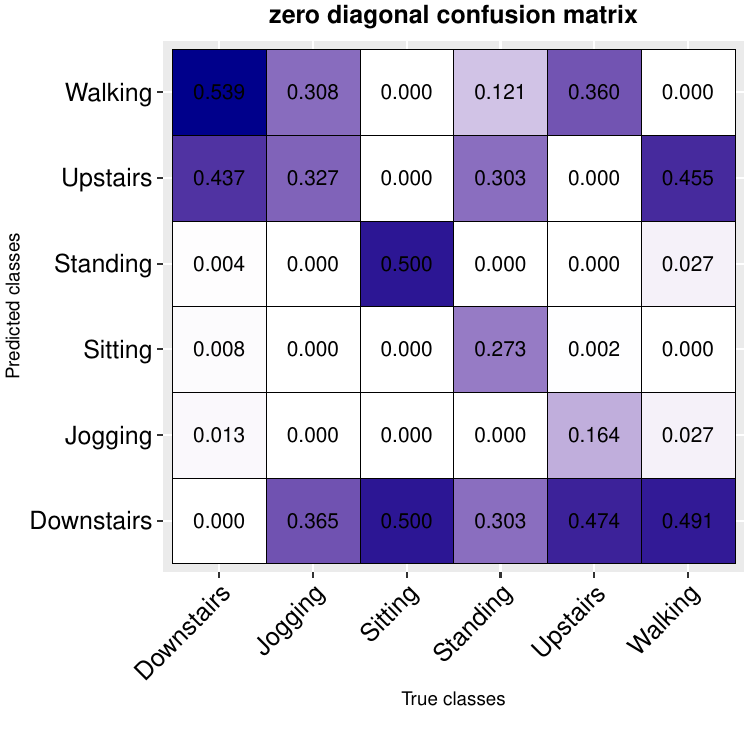}\hfill
\includegraphics[width=.40\textwidth]{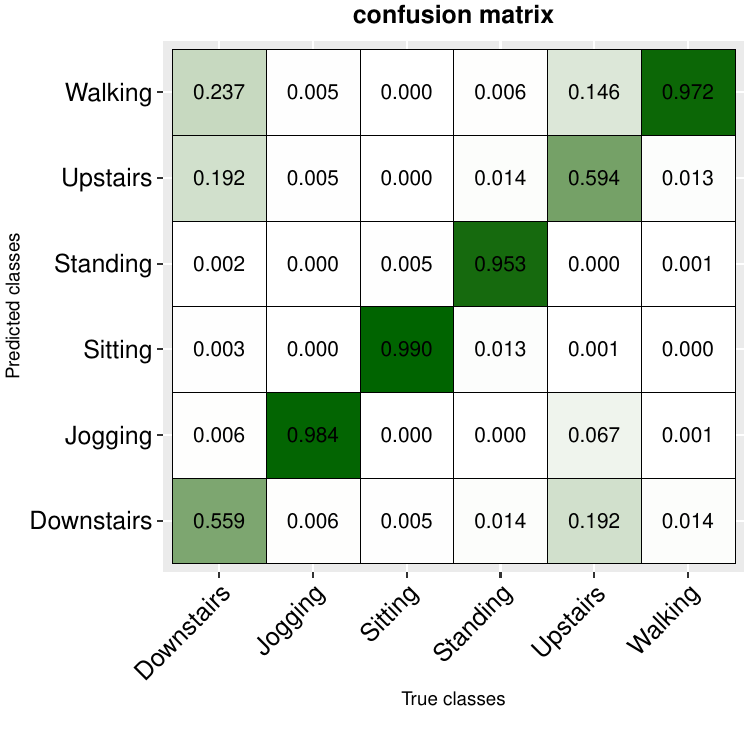}\hfill

\includegraphics[width=.60\textwidth]{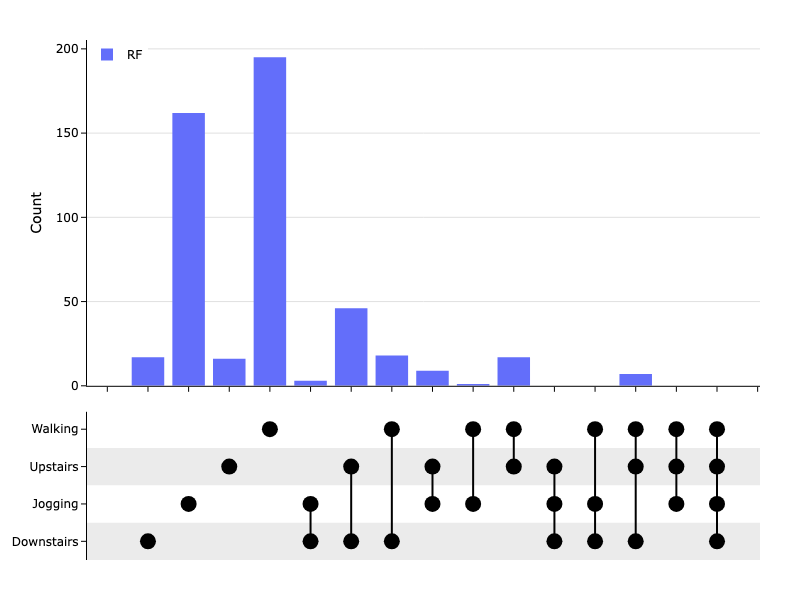}\hfill

\caption{D1: WISDM. Co-occurrence matrix (upper-left), co-occurrence graph (upper-right), zero diagonal confusion matrix (middle-left), confusion matrix (middle-right), and multiset-enumerating bar chart (bottom-center) for mixed model with RF.}
\label{fig:d1-plots-rf-mm}
\end{figure}

Figure~\ref{fig:d1-plots-rf-mm} shows the co-occurrence plot and the co-occurrence graph at the top, the zero diagonal confusion matrix (ZDCM) and the confusion matrix in the middle, and the multiset-enumerating chart at the bottom (for the mixed model with RF). The top plots capture information about the resulting conformal prediction sets. From the co-occurrence matrix, it can be seen that the classes with which \emph{`Downstairs'} co-occurred the most were \emph{`Walking'} ($33.5\%$) and \emph{`Upstairs'} ($62.9\%$). \emph{`Sitting'} co-occurred the most with \emph{`Standing'} which makes sense since both are low-acceleration activities. From the co-occurrence graph it can be seen that the pair \emph{`Downstairs'}-\emph{`Upstairs'} is the one that occurred the most (thickest edge). From the smallest nodes sizes, we see that \emph{`Standing'} and \emph{`Sitting'} are the classes that appeared less frequently together with other classes.

The middle plots capture information about the single predictions of the non-conformal models. The ZDCM shows that for example, \emph{`Jogging'} was confused many times with \emph{`Walking'} and \emph{`Upstairs'}. It is also worth noting the relationship between graphs. The co-occurrence matrix and the ZDCM look very similar. That is, if two classes often appear together in the prediction set of the conformal classifier, it is also likely that they will be confused by the non-conformal classifier. Similarly, in the co-occurrence graph the biggest nodes correspond to \emph{`Downstairs'}, \emph{`Upstairs'}, and \emph{`Walking'} which are also the most colored ones (rows) in the corresponding confusion matrix. From the multiset-enumerating chart we can see that apart from the singletons, the most frequent prediction sets were the ones containing the set \{\emph{`Upstairs'},\emph{`Downstairs'}\}.

\begin{figure}[h!]
\centering
\includegraphics[width=0.4\textwidth]{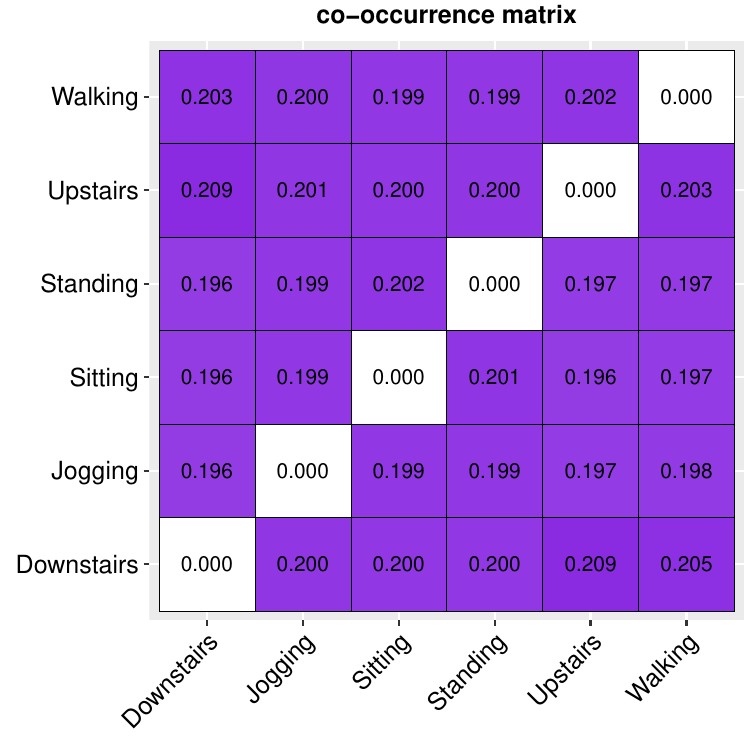}
\caption{Co-occurrence matrix for KNN.}\label{fig:d1-com-knn-mm}
\end{figure}

The visual patterns in the figures looked similar for the other types of models and classifiers (except KNN). For clarity, here we only include plots for the mixed model and RF but the published source code generates all of them. For KNN, the co-occurrence plot looks uniform (Figure~\ref{fig:d1-com-knn-mm}). As previously mentioned, KNN is also the classifier with the biggest set sizes. This is because KNN does not produce very well calibrated prediction scores. It's prediction scores are computed as the fraction of nearest neighbors that belong to the true class. For example, if $k=3$ and the true class is \emph{`A'}, and all the nearest neighbors are of class \emph{`A'}, then the score for \emph{`A'} will be $100\%$. The same behavior was observed for all the remaining datasets.

\begin{figure}[h!]
\centering
\includegraphics[width=0.7\textwidth]{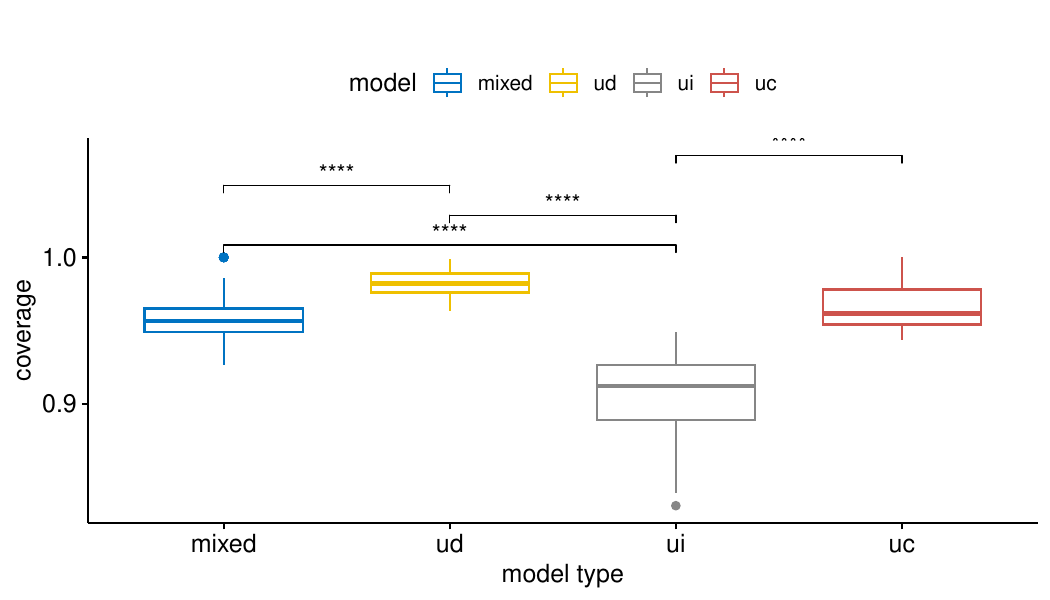}
\caption{D1: WISDM. Boxplots of coverage between models. ns: $p > 0.05$, *: $p <= 0.05$, **: $p <= 0.01$, ***: $p <= 0.001$, ****: $p <= 0.0001$. }\label{fig:d1-boxplot-coverage}
\end{figure}

Figure~\ref{fig:d1-boxplot-coverage} shows boxplots of the coverage for all model types across all classifiers. In order to test the hypotheses defined in Section~\ref{sec:introduction},  the significance levels between models are also shown (t-test). Here, we can see that all comparisons are statistically significant and thus, we reject the three hypotheses. Similar to non-conformal models, the conformal UDM was the best performing one. The UCM performed better than the UIM.

\begin{figure}[h!]
\centering
\includegraphics[width=0.7\textwidth]{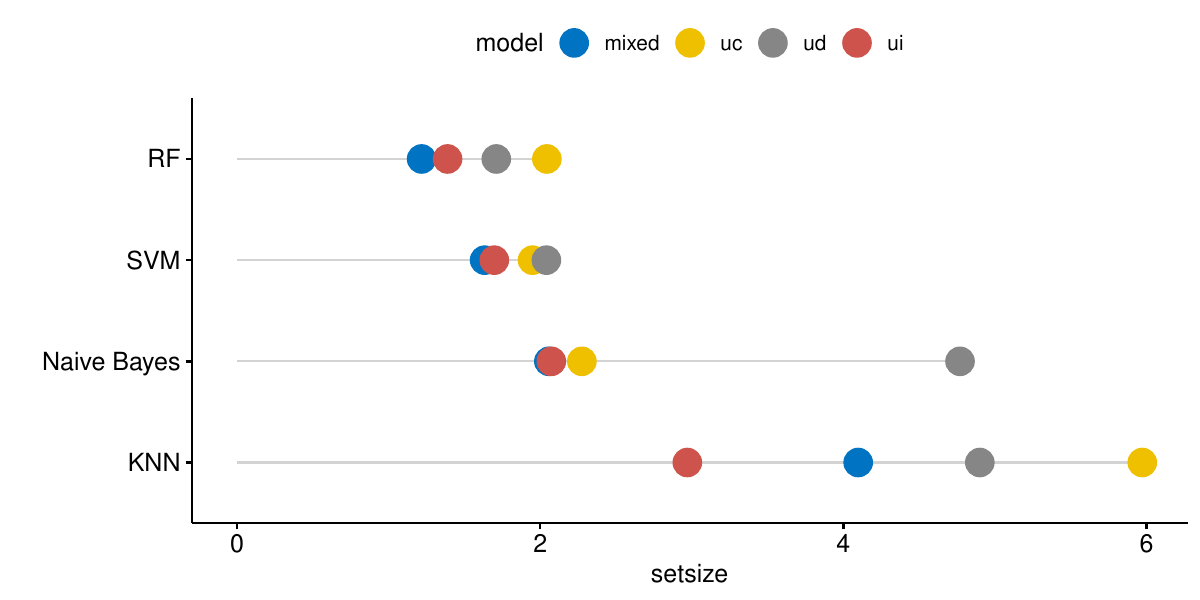}
\caption{D1: WISDM. Lolliplot of set size for all classifiers and models.}\label{fig:d1-lolliplot-setsize}
\end{figure}

Figure~\ref{fig:d1-lolliplot-setsize} shows a plot comparing the set size between classifiers and for all models. Here we can see that RF achieves the smallest sets followed by SVM and Naive Bayes. As already seen, KNN produces the biggest sets.

\begin{figure}[h!]
\centering
\includegraphics[width=.40\textwidth]{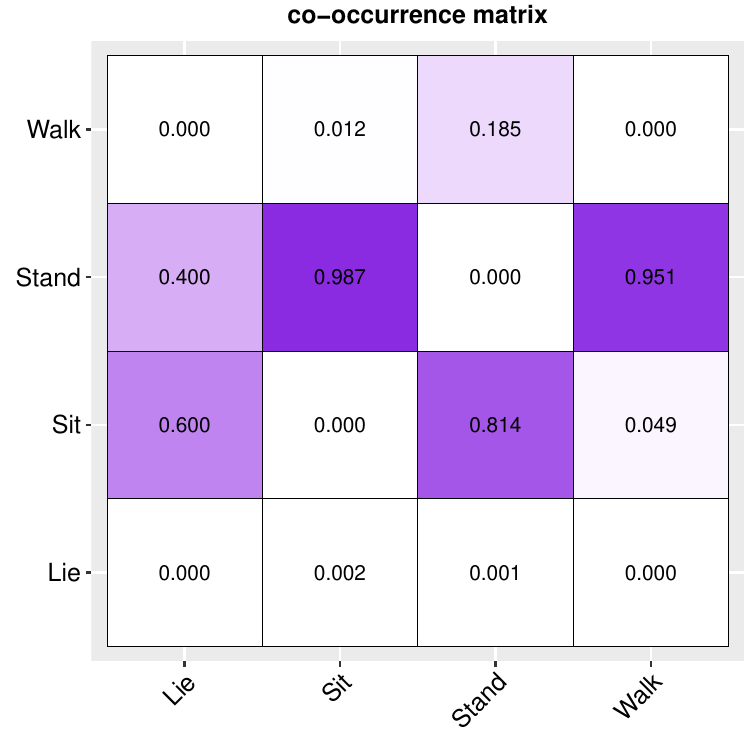}\hfill
\includegraphics[width=.40\textwidth]{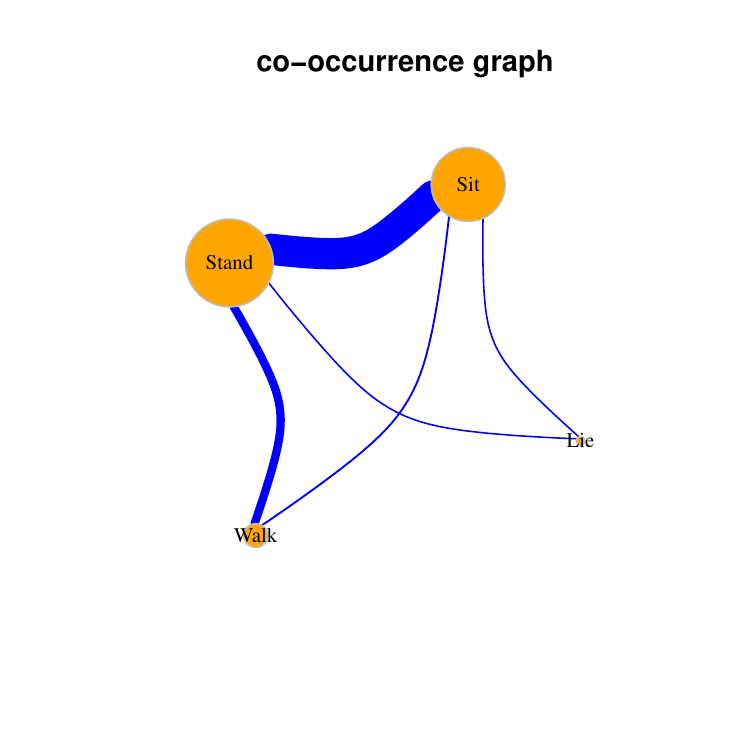}\hfill

\includegraphics[width=.40\textwidth]{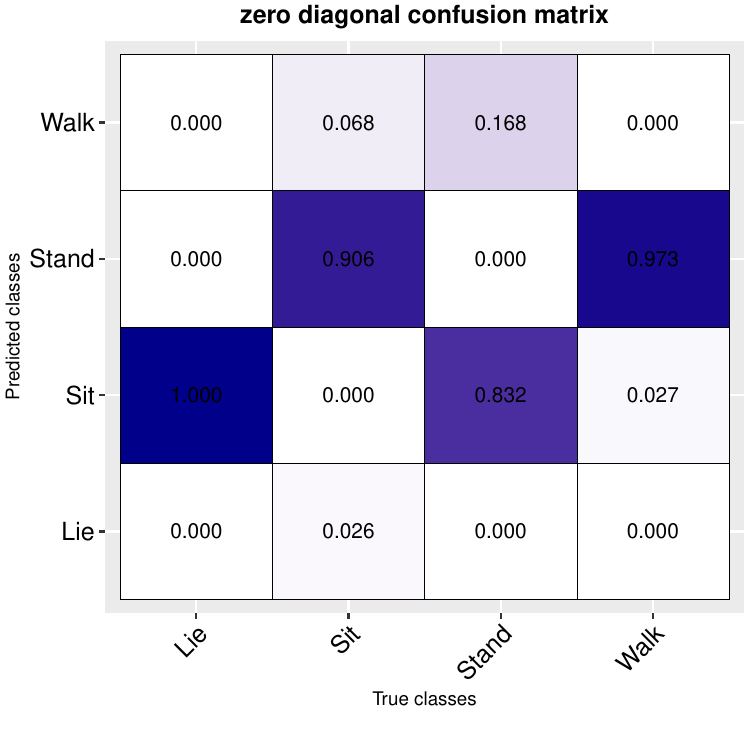}\hfill
\includegraphics[width=.40\textwidth]{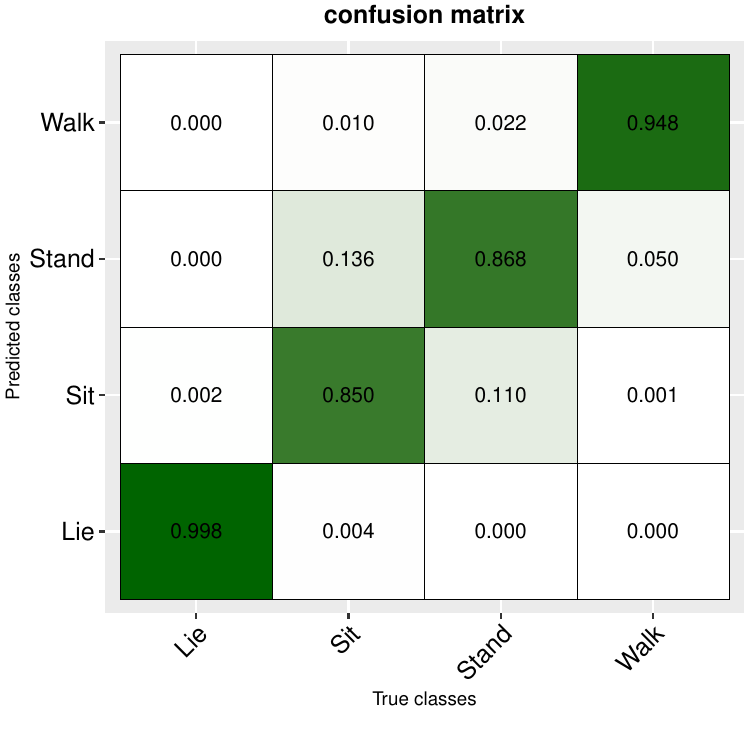}\hfill

\includegraphics[width=.60\textwidth]{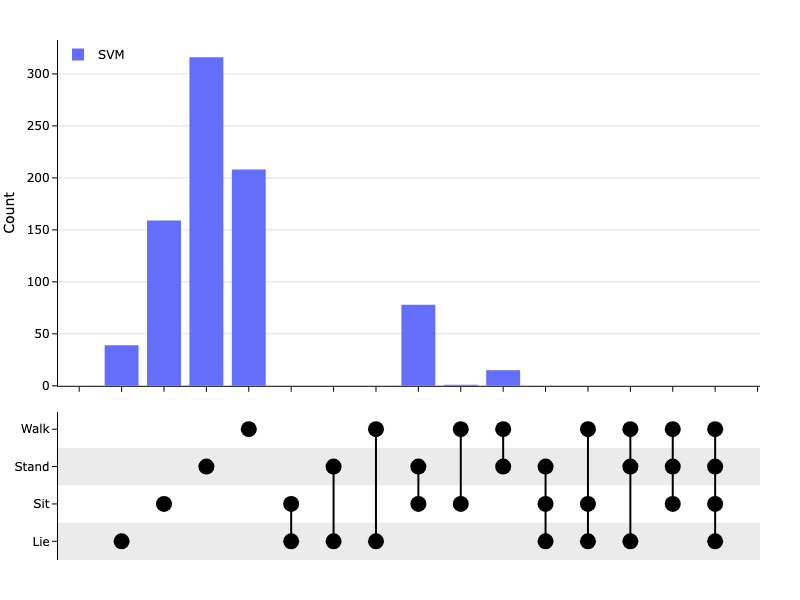}\hfill

\caption{D2: Opportunity. Co-occurrence matrix (upper-left), co-occurrence graph (upper-right), zero diagonal confusion matrix (middle-left), confusion matrix (middle-right), and multiset-enumerating bar chart (bottom-center) for mixed model with SVM.}
\label{fig:d2-plots-svm-mm}
\end{figure}

Figure~\ref{fig:d2-plots-svm-mm} shows the resulting plots for SVM and dataset D2. Here, we can see similar patterns than those for D1 using RF (Figure~\ref{fig:d1-plots-rf-mm}).

\begin{figure}[h!]
\centering
\includegraphics[width=0.7\textwidth]{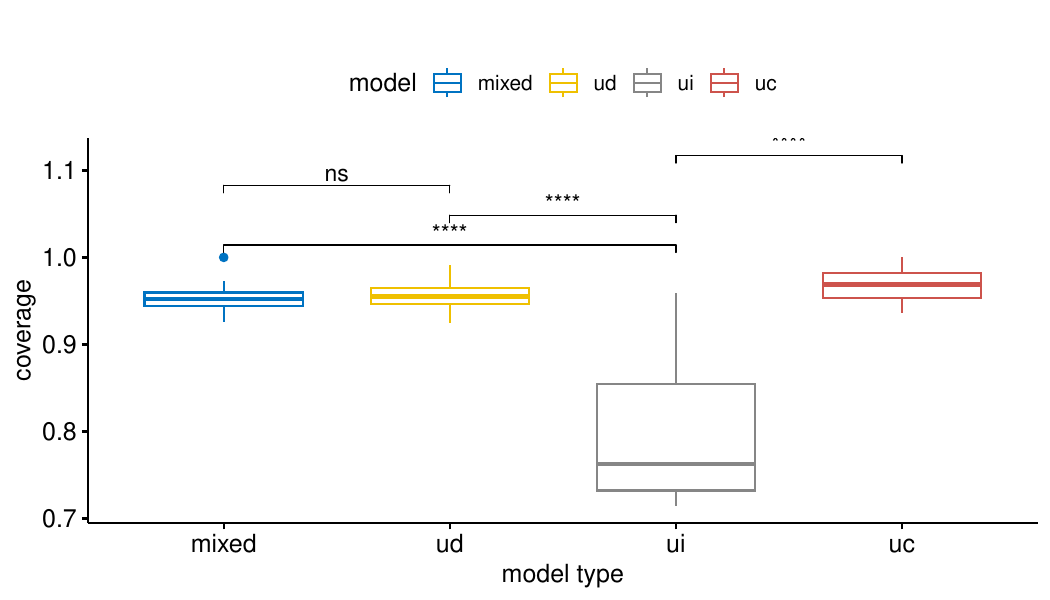}
\caption{D2: Opportunity. Boxplots of coverage between models. ns: $p > 0.05$, *: $p <= 0.05$, **: $p <= 0.01$, ***: $p <= 0.001$, ****: $p <= 0.0001$. }\label{fig:d2-boxplot-coverage}
\end{figure}

\begin{figure}[h!]
\centering
\includegraphics[width=0.7\textwidth]{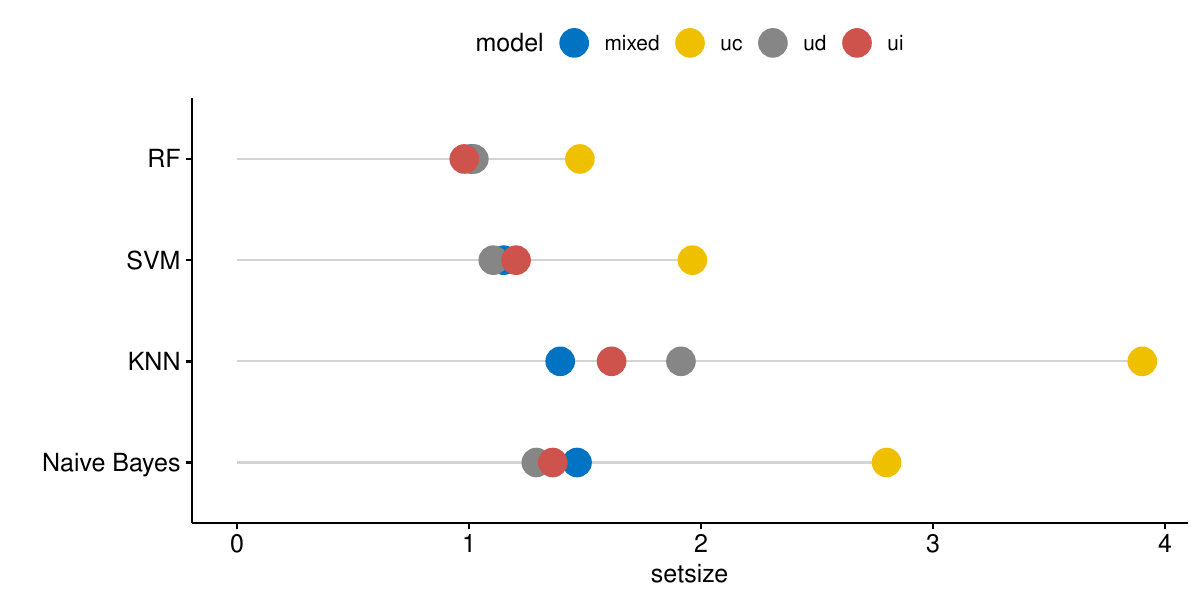}
\caption{D2: Opportunity. Lolliplot of set size for all classifiers and models.}\label{fig:d2-lolliplot-setsize}
\end{figure}

Figure~\ref{fig:d2-boxplot-coverage} shows the coverage boxplots for dataset D2. In this case the difference between the mixed and the UDM was not significant. Figure~\ref{fig:d2-lolliplot-setsize} had a similar pattern compared to dataset D1 being RF and SVM the classifiers that produced the smallest sets.

\begin{figure}[h!]
\centering
\includegraphics[width=0.7\textwidth]{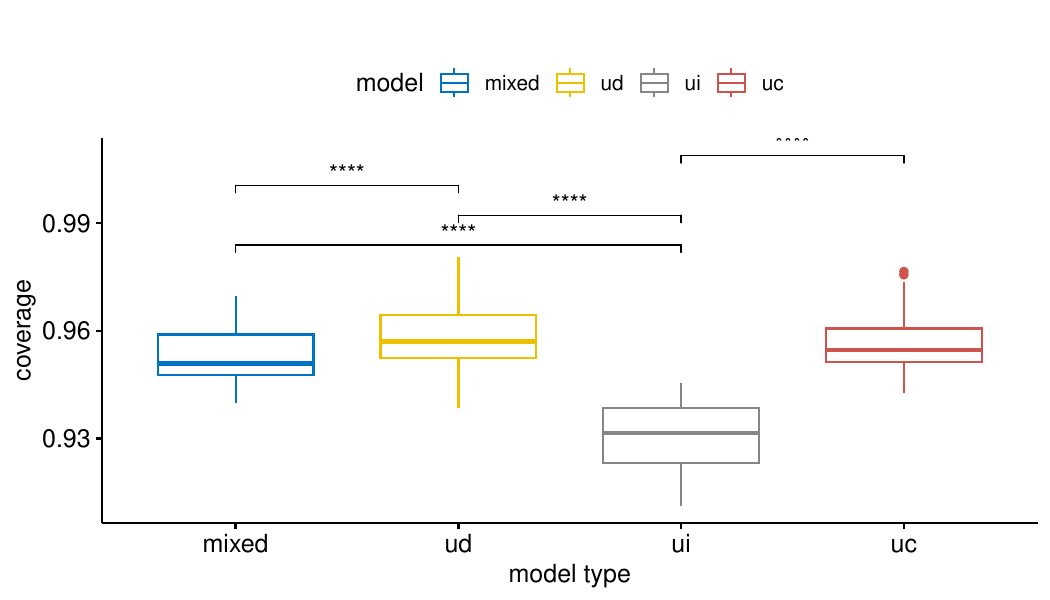}
\caption{D3: HAR70+. Boxplots of coverage between models. ns: $p > 0.05$, *: $p <= 0.05$, **: $p <= 0.01$, ***: $p <= 0.001$, ****: $p <= 0.0001$. }\label{fig:d3-boxplot-coverage}
\end{figure}

For dataset D3, the patterns in the plots look similar, thus, they were not included here. Figure~\ref{fig:d3-boxplot-coverage} shows the resulting coverage boxplots. In this case, the differences between the compared models were all significant.

\begin{figure}[h!]
\centering
\includegraphics[width=0.7\textwidth]{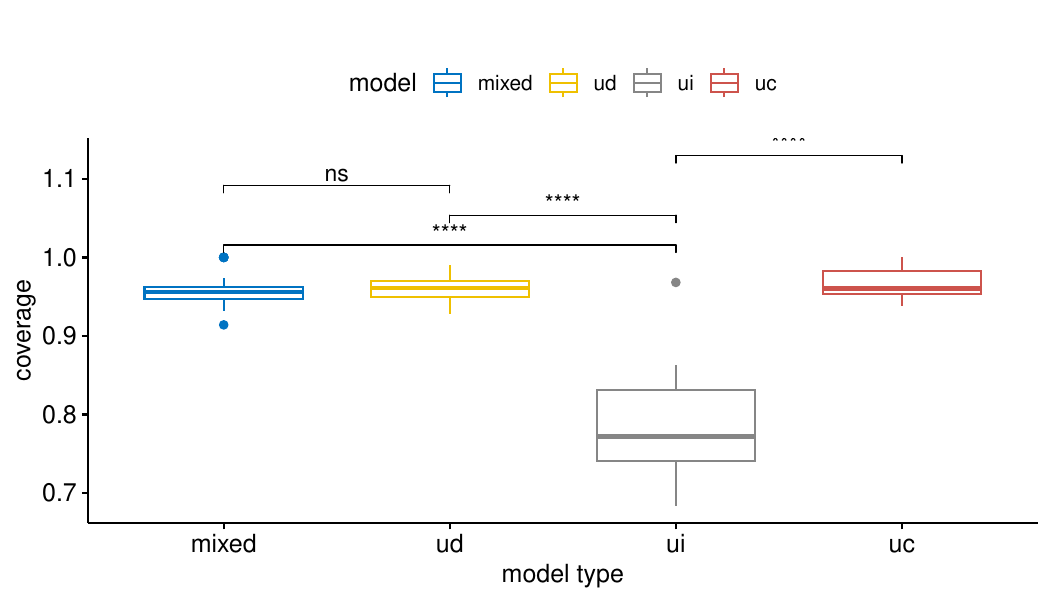}
\caption{D4: Smartwatch Gestures. Boxplots of coverage between models. ns: $p > 0.05$, *: $p <= 0.05$, **: $p <= 0.01$, ***: $p <= 0.001$, ****: $p <= 0.0001$. }\label{fig:d4-boxplot-coverage}
\end{figure}

Similar to dataset D2, the coverage difference between the mixed and the UDM was not significant for dataset D4 (Figure~\ref{fig:d4-boxplot-coverage}).

\section{Conclusion}\label{conslusions}

In this work we conducted exhaustive tests of several evaluation strategies in multi-user systems within the framework of conformal prediction. We reported conformal and non-conformal performance measures and proposed several prediction set visualizations to better understand the results of conformal prediction sets and how they relate to non-conformal (singleton) predictions. We also proposed an evaluation strategy (UCM) where first, a general model is trained and then, it is calibrated using data from the target user. 

From the results, we observed that RF produced the smallest sets while also attaining the highest performance in terms of non-conformal measures which is in agreement with previous work \citep{jourdan_neurips_2021}. Overall, and as expected, the UIM obtained the lowest performance in all cases which aligns with the results of previous works. For all datasets, the MM was better than the UIM which can lead to overestimation of the performance of a system since as previously mentioned, the MM approach is the most common one used in the literature to evaluate a system, however, it does not take into account the inter-user variance. This was the case for both conformal and non-conformal measures.

The three stated null hypotheses were rejected for all datasets except hypothesis 2 since there was no significant difference between the UDP and the MM for 2 out of the 4 datasets. For all datasets, the co-occurrence matrix plot looked similar to the zero diagonal confusion matrix and the co-occurrence graph biggest nodes corresponded with the most row-colored classes in the confusion matrix.

In terms of coverage, the UCM achieved better performance than the UIM despite using less calibration data. This shows that for multi-user systems, more data doesn't necessarily mean better. This is because the data may not have been generated by the same distribution.

In this work we have focused on 4 multi-user evaluation strategies. However, there is still another one, often called user-adaptive model or UAM for short~\citep{garcia2020adaptive}. A UAM tries to mitigate the weaknesses of a UIM and a UDM. That is, it tries to achieve better performance than a general model while trying to minimize the training data from the target user. In future work we plan to evaluate this types of models that adapt over time within conformal prediction. We also plan to further extend the analysis of the proposed visualization methods for prediction sets.



\backmatter

\bmhead{Supplementary information}

The source code to reproduce the results is available here: \url{https://github.com/enriquegit/conformal-prediction-multiuser}.



\section*{Declarations}

\textbf{Authors' contributions} Study conception and design by EGC. Datasets collection and preprocessing by EGC and NJ. Visualizations by EGC and LGB. All authors contributed to the analysis of results and document writing.  All authors read and approved the final manuscript.

\

\noindent \textbf{Competing interests} The authors have no competing interests to declare that are relevant to the content of this article.

\

\noindent \textbf{Funding} No funding was received for conducting this study.

\

\noindent \textbf{Data availability} All the datasets used in this study are open access. The corresponding references for each dataset are listed in section~\ref{sec:datasets}.

\

\noindent \textbf{Code availability} The code and instructions on how to reproduce all the results in this study are available here: \url{https://github.com/enriquegit/conformal-prediction-multiuser}.

\

\noindent \textbf{Ethics approval} Not applicable.

\

\noindent \textbf{Consent to participate} Not applicable.

\

\noindent \textbf{Consent for publication} Not applicable.

\









\bibliography{sn-bibliography}

\end{document}